\documentclass{article}

\usepackage[numbers]{natbib}

\usepackage[final]{nips_2018}

\usepackage[utf8]{inputenc} 
\usepackage[T1]{fontenc}    
\usepackage{hyperref}       
\usepackage{url}            
\usepackage{booktabs}       
\usepackage{amsfonts}       
\usepackage{nicefrac}       
\usepackage{microtype}      
\usepackage{algorithm}
\usepackage{algorithmicx}
\usepackage{algpseudocode}
\usepackage{eqparbox}
\usepackage{amsmath}
\usepackage{wrapfig}
\usepackage{subfigure}
\usepackage{graphicx}
\usepackage{float}
\usepackage{bbm}
\usepackage{multirow}
\usepackage{amssymb}
\usepackage[
  separate-uncertainty = true,
  multi-part-units = repeat
]{siunitx}

\title{Learning to Teach with Dynamic Loss Functions}

\author{
  $^1$Lijun Wu$^{\dagger}$\thanks{The work was done when the first and fourth authors were interns at Microsoft Research Asia.},  $^2$Fei Tian\thanks{The first two authors contribute equally to this work.},  $^2$Yingce Xia,  $^3$Yang Fan$^{\star}$, \\
  \textbf{$^2$Tao Qin,  $^1$Jianhuang Lai,  $^2$Tie-Yan Liu}\\
  $^1$Sun Yat-sen University, Guangzhou, China\quad$^2$Microsoft Research, Beijing, China\\
  $^3$University of Science and Technology of China, Hefei, China\\
  $^1$wulijun3@mail2.sysu.edu.cn,\; stsljh@mail.sysu.edu.cn\\
  $^2$\{fetia, yingce.xia, taoqin, tie-yan.liu\}@microsoft.com,\; $^3$fyabc@mail.ustc.edu.cn
}

\begin{document}

\maketitle

\begin{abstract}
Teaching is critical to human society: it is with teaching that prospective students are educated and human civilization can be inherited and advanced. A good teacher not only provides his/her students with qualified teaching materials (e.g., textbooks), but also sets up appropriate learning objectives (e.g., course projects and exams) considering different situations of a student. When it comes to artificial intelligence, treating machine learning models as students, the loss functions that are optimized act as perfect counterparts of the learning objective set by the teacher. In this work, we explore the possibility of imitating human teaching behaviors by dynamically and automatically outputting appropriate loss functions to train machine learning models. Different from typical learning settings in which the loss function of a machine learning model is predefined and fixed, in our framework, the loss function of a machine learning model (we call it student) is defined by another machine learning model (we call it teacher). The ultimate goal of teacher model is cultivating the student to have better performance measured on development dataset. Towards that end, similar to human teaching, the teacher, a parametric model, dynamically outputs different loss functions that will be used and optimized by its student model at different training stages. We develop an efficient learning method for the teacher model that makes gradient based optimization possible, exempt of the ineffective solutions such as policy optimization. We name our method as ``learning to teach with dynamic loss functions'' (L2T-DLF for short). Extensive experiments on real world tasks including image classification and neural machine translation demonstrate that our method significantly improves the quality of various student models. 
\end{abstract}


\section{Introduction}

Teaching, which aims to help students learn new knowledge or skills effectively and efficiently, is important to advance modern human civilization. In human society, the rapid growth of qualified students not only relies on their intrinsic learning capability, but also, even more importantly, relies on the substantial guidance from their teachers. The duties of teachers cover a wide spectrum: defining the scope of learning (e.g., the knowledge and skills that we expect students to demonstrate by the end of a course), choosing appropriate instructional materials (e.g., textbooks), and assessing the progress of students (e.g., through course projects or exams). Effective teaching involves progressively and dynamically refining the teaching strategy based on reflection and feedback from students. 

Recently, the concept of teaching has been introduced into artificial intelligence (AI), so as to improve the learning process of a machine learning model. Currently, teaching in AI mainly focuses on training data selection. For example, \emph{machine teaching}~\cite{machine_teaching, teaching+dimension, iterative_teaching} aims at identifying the smallest training data that is capable of producing the optimal learner models. The very recent work, \emph{learning to teach} (L2T for short)~\cite{l2t_iclr}, demonstrates how to automatically design teacher models for better machine learning process. While conceptually L2T can cover different aspects of teaching in AI, \cite{l2t_iclr} only studies the problem of training data teaching.

In this work, inspired from learning to teach, we study loss function teaching in a formal and concrete manner for the first time. The main motivation of our work is a natural observation on the analogy between loss functions in machine learning and exams in educating human students: appropriate exams reflect the progress of students and urge them to make improvements accordingly, while loss values outputted by the loss function evaluate the performance of current machine learning model and set the optimization direction for the model parameters. 

\begin{wrapfigure}[]{}{0.35\textwidth}
\centering
\includegraphics[width=\linewidth]{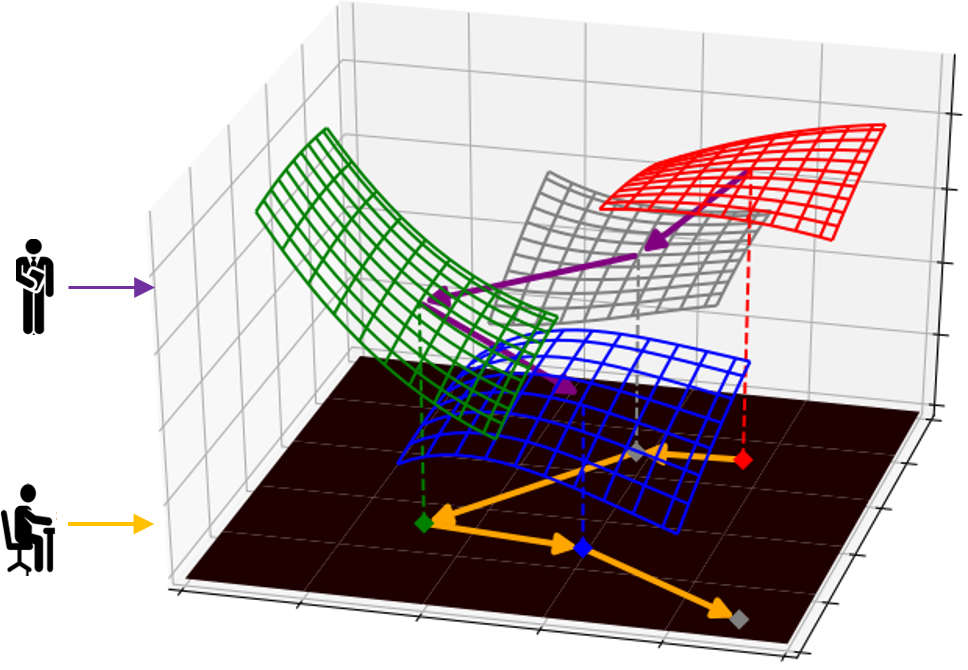}
\caption{The student model is trained via minimizing the dynamic loss functions taught by the teacher model (yellow curve). The bottom black plane represents the parameter space of student model and the four colored mesh surfaces denote different loss functions outputted via teacher model at different phases of student model training.}
\label{fig:dynamic}
\end{wrapfigure}

In our loss function teaching framework, a teacher model plays the role of outputting loss functions for the student model (i.e., the daily machine learning model to solve a task) to minimize. Inspired from human teaching, we design the teacher model according to the following principles. First, similar to the different difficulty levels of exams with respect to the progress of student in human education, the loss function set by the teacher model should be \emph{dynamic}, i.e., the loss functions should be adaptive to different phases of the training process of the student model. To achieve this, we require our teacher model to take the status of student model into consideration in setting the loss functions, and to dynamically change the loss functions with respect to the growth of the student model. Such process is shown in Fig. \ref{fig:dynamic}. Second, the teacher model should be able to make self-improvement, just as a human teacher can accumulate more knowledge and improve his/her teaching skills through more teaching practices. To achieve that, we assume the loss function takes the form of neural network whose coefficients are determined via a parametric teacher model, which is also a neural network. The parameters of the teacher model can be automatically optimized in the teaching process. Through optimization, the teacher keeps improving its teaching model and consequently the quality of loss functions it outputs. We name our method as \emph{learning to teach with dynamic loss functions} (L2T-DLF).

The eventual goal of the teacher model is that its output can serve as the loss function of the student model to maximize the long-term performance of the student, measured via a task-specific objective such as 0-1 accuracy in classification and BLEU score in sequence prediction~\cite{BLEU}, on a stand-alone development dataset. Learning a good teaching model is not trivial, since on the one hand the task-specific objective is usually non-smooth w.r.t. student model outputs, and on the other hand the final evaluation of the student model is incurred on the dev set, disjoint with the training dataset where the teaching process actually happens. We design an efficient gradient based optimization algorithm to optimize teacher models. Specifically, to tackle the first challenge, we smooth the task-specific measure to its expected version where the expectation is taken on the direct output of student model. To address the second challenge, inspired by Reverse-Mode Differentiation (RMD)~\cite{autoDiff,RMD,RMD_rev}, through reversing the stochastic gradient descent training process of the student model, we obtain derivatives of the parameters of the teacher model via chaining backwards the error signals incurred on the development dataset . 

We demonstrate the effectiveness of L2T-DLF on various real-world tasks including image classification and neural machine translation with different student models such as multi-layer perception networks, convolutional neural networks and sequence-to-sequence models with attention. The improvements clearly demonstrate the effectiveness of the new loss function learnt by L2T-DLF.

\section{Related Work}
The study of teaching for AI, inspired by human teaching process, has a long history~\cite{teach_old_1,teach_old_2}. The most recent efforts of teaching mainly focus on the level of training data selection. For example, the \emph{machine teaching}~\cite{machine_teaching, teaching+dimension, iterative_teaching} literature targets at building the smallest training set to obtain a pre-given optimal student model. A teaching strategy is designed in~\cite{easy_l2t_0, easy_l2t_1} to iteratively select unlabeled data to label within the context of multi label propagation, in a similar manner with curriculum learning~\cite{CL, SPL}.  Furthermore there are research on \emph{pedagogical teaching} inspired from cognitive science~\cite{Pedagogical, ShowingDemo,openAIpeda} in which a teacher module is responsible for providing informative examples to the learner for the sake of understanding a concept rapidly.

The recent work \emph{learning to teach} (L2T)~\cite{l2t_iclr} offers a more comprehensive view of teaching for AI, including training data teaching, loss function teaching and hypothesis space teaching. Furthermore, L2T breaks the strong assumption towards the existence of an optimal off-the-shelf student model adopted by previous \emph{machine teaching} literature~\cite{machine_teaching, iterative_teaching}. Our work belongs to the general framework of L2T, with a particular focus on a thorough landscape of loss function teaching, including the detailed problem setup and efficient solution for dynamically setting loss functions for training machine learning models.

Our work, and the more general L2T, leverages automatic techniques to bypass human prior knowledge as much as possible, which is in line with the principles of \emph{learning to learn} and \emph{meta learning}~\cite{metaLearning,L2LBook,learning2learn_sgd, NAS,NAO, learning_to_optimize,l2l_wo_gd,MAML}. What makes our work different with others, from the technical point of view, is that: 1) we leverage gradient based optimization method rather than reinforcement learning~\cite{NAS,l2t_iclr}; 2) we need to handle the difficulty when the error information cannot be directly back propagated from the loss function, since we aim at discovering the best loss function for the machine learning models. We design an algorithm based on  Reverse-Mode Differentiation (RMD)~\cite{RMD,RMD_rev,forwardRMD} to tackle such a difficulty.


Specially designed loss functions play important roles in boosting the performances of real-world tasks, either by approximating the non-smooth task-specific objective such as 0-1 accuracy in classification~\cite{smooth_acc}, NDCG in ranking~\cite{softrank}, BLEU in machine translation~\cite{rl4nmt, ac_s2s} and MAP in object detection~\cite{henderson2016end, direct_opt}, or easing the optimization process of the student model such as overcoming the difficulty brought by data imbalance ~\cite{uneven_margin,FocalLoss} and numerous local optima~\cite{grad_opt}. L2T-DLF differs from prior works in that: 1) the loss functions are automatically learned, covering a large space and without the demand of heuristic understanding for task specific objective and optimization process; 2) the loss function dynamically evolves during the training process, leading to a more coherent interaction between loss and student model.

\section{Model}

In this section, we introduce the details of L2T-DLF, including the student model and the teacher model, as well as the training strategy for optimizing the teacher model.

\subsection{Student Model}

For a task of interest, we denote its input space and output space respectively as $\mathcal{X}$ and $\mathcal{Y}$. The student model for this task is then denoted as $f_\omega: \mathcal{X}\rightarrow \mathcal{Y}$, with $\omega$ as its weight parameters. The training of student model $f_\omega$ is an optimization process that discovers a good weight parameter $\omega^*$ within a hypothesis space $\Omega$, by minimizing a loss function $l$ on the training data $D_{train}$ containing $M$ data points $D_{train}=\{(x_i,y_i)\}_{i=1}^M$. Specifically $\omega^*$ is obtained via solving $\min_{\omega\in \Omega}\sum_{(x,y)\in D_{train}}l(f_\omega(x),y)$. For the convenience of description, we define a new notation $L(f_\omega, D)=\sum_{(x,y)\in D}l(f_\omega(x),y)$ where $D$ is a dataset and will simultaneously name $L$ as loss function when the context is clear. The learnt student model $f_{\omega^*}$ is then evaluated on a test data set $D_{test}=\{(x_i,y_i)\}_{i=1}^N$ to obtain a score $\mathcal{M}(f_{\omega^*},D_{test})=\sum_{(x,y)\in D_{test}}m(f_{\omega^*}(x),y)$, as its performance. Here the task specific objective $m(y_1, y_2)$ measures the similarity between two output candidates $y_1$ and $y_2$. 

The loss function $l(\hat{y},y)$, taking the model prediction $\hat{y}=f_\omega(x)$ and ground-truth $y$ as inputs, acts as the surrogate of $m$ to evaluate the student model $f_\omega$ during its training process, just as the exams in real-world human teaching. We assume $l(\hat{y},y)$ is a neural network with some coefficients $\Phi$, denoted as $l_\Phi(\hat{y},y)$. It can be a simple linear model, or a deep neural network (some concrete examples are provided in section~\ref{subsec:exp_image} and section~\ref{subsec:exp_nmt}). With such a loss function $l_\Phi(\hat{y},y)$ (and the induced notation $L_\Phi$), the student model gets sequentially updated via minimizing the output value of $l_\Phi$ by, for example, stochastic gradient descent (SGD): $\omega_{t+1} = \omega_t-\eta_t \frac{\partial L_{\Phi}(f_{\omega_t}, D_{train}^t)}{\partial \omega_t}, t=\{1,2,\cdots, T\}$, where $D_{train}^t\subseteq D_{train}$, $\omega_t$ and $\eta_t$ is respectively the mini-batch training data, student model weight parameter and learning rate at $t$-th timestep. For ease of statement we simply set $\omega^*=\omega_T$. 


\subsection{Teacher Model}
A teacher model is responsible for setting the proper loss function $l$ to the student model by outputting appropriate loss function coefficients $\Phi$. To cater for different status of student model training, we ask the teacher model to output different loss functions $l^{t}$ at each training step $t$. To achieve that, the status of a student model is represented by a state vector $s_t$ at timestep $t$, which contains for example the current training/dev accuracy and iteration number. The teacher model, denoted as $\mu$, then takes $s_t$ as inputs to compute the coefficients of loss function $\Phi_t$ at $t$-th timestep as $\Phi_t = \mu_\theta(s_t)$, where $\theta$ is the parameters of the teacher model. We further provide some examples of $\mu_\theta$ in section~\ref{subsec:exp_image} and section~\ref{subsec:exp_nmt}. The actual loss function for student model is then $l^t=l_{\Phi_t}$. The learning process of student model then switches to:
\begin{equation}
\label{eqn:sgd_seq}
\omega_{t+1} = \omega_t - \eta_t \frac{\partial L_{\Phi_t}(f_{\omega_t}, D_{train}^t)}{\partial \omega_t}=\omega_t-\eta_t\frac{\partial L_{\mu_{\theta}(s_t)}(f_{\omega_t}, D_{train}^t)}{\partial \omega_t}.
\end{equation}
Such a sequential procedure of obtaining $f_{\omega^*}$ (i.e., $f_{\omega_T}$) is the learning process of the student model with training data $D_{train}$ and loss function provided via the teacher model $\mu_\theta$, and we use an abstract operator $\mathcal{F}$ to denote it: $f_{\omega^*}=\mathcal{F}(D_{train}, \mu_\theta)$.  

Just as the training and testing setup in typical machine learning scenarios, the teacher model here similarly follows the two phases setup. Specifically, in the training process of teacher model, similar to qualified human teachers are good at improving the quality of exams,  the teacher model in L2T-DLF refines the loss function it sets up via optimizing its own $\theta$. The ultimate goal of teacher model is to maximize the performance of induced student model on a stand-alone development dataset $D_{dev}$:
\begin{equation}
\label{eqn:goal_teacher}
\max_\theta \mathcal{M}(f_{\omega^*}, D_{dev})=\max_\theta \mathcal{M}(\mathcal{F}(D_{train}, \mu_\theta), D_{dev}).
\end{equation}
We introduce the detailed training process (i.e., how to efficiently optimize Eqn. (\ref{eqn:goal_teacher})) in section~\ref{subsec:train_teacher}. In the testing process of the teacher model, $\theta$ is fixed and the student model $f_\omega$ gets updated with the guidance of teacher model $\mu_{\theta}$, as specified in Eqn. (\ref{eqn:sgd_seq}).

\subsection{Training Process of Teacher Model}
\label{subsec:train_teacher}

There are two challenges to optimize teacher model: 1) the evaluation measure $m$ is typically non-smooth and non-differentiable w.r.t. the parameters of student model; 2) the error is incurred on dev set while the teacher model plays effect in training phase.

We use continuous relaxation of $m$ to tackle the first challenge. The main idea is to inject randomness into $m$ to form an approximated version $\tilde{m}$, where the randomness comes from the student model~\cite{softrank}. Thanks to the fact that quite a few student models output probabilistic distributions on $\mathcal{Y}$, the randomness naturally comes from the direct outputs of $f_\omega$. Specifically, to approximate the performance of $f_\omega$ on a test data sample $(x,y)$, we have $\tilde{m}(f_\omega(x), y)=\sum_{y^*\in \mathcal{Y}}m(y^*, y)p_\omega(y^*|x)$, where $p_\omega(y^*|x)$ is the probability of predicting $y^*$ given $x$ using $f_\omega$. The gradient of $\omega$ is then easy to obtain via $\frac{\partial \tilde{m}(f_\omega(x),y)}{\partial \omega}=\sum_{y^*\in \mathcal{Y}}m(y^*,y)\frac{\partial p_\omega(y^*|x)}{\partial\omega}$. We further introduce a new notation $\tilde{\mathcal{M}}(f_\omega, D_{dev})=\sum_{(x,y)\in D_{dev}}\tilde{m}(f_\omega(x),y)$ which approximates the objective of the teacher model $\mathcal{M}(f_{\omega_T},D_{dev})$.

We use Reverse-Mode Differentiation (RMD)~\cite{autoDiff,RMD, RMD_rev} to fill in the gap between training data and development data. To better show the RMD process, we can view the sequential process in Eqn. (\ref{eqn:sgd_seq})  as a special feed-forward process of a deep neural network where each $t$ corresponds to one layer, and RMD corresponds to the backpropagation process looping the SGD process backwards from $T$ to $1$. Specifically denote $d\theta$ as the gradient of $\tilde{M}(f_{\omega_T}, D_{dev})$ w.r.t. the teacher model parameters $\theta$, which has initial value $d\theta=0$. On the dev dataset $D_{dev}$, the gradient of $\tilde{\mathcal{M}}(f_\omega, D_{dev})$ w.r.t. the parameter of student model $\omega_T$ is calculated as 
\begin{equation}
\label{eqn:dOmegaT}
d\omega_T=\frac{\partial \tilde{\mathcal{M}}(f_{\omega_T},D_{dev})}{\partial \omega_T}=\sum_{(x,y)\in D_{dev}}\frac{\partial \tilde{m}(f_{\omega_T}(x),y)}{\partial \omega_T}.
\end{equation}
Then looping backwards from $T$ and corresponding to Eqn. (\ref{eqn:sgd_seq}), at each step $t=\{T-1,\cdots, 1\}$ we have
\begin{equation}
\label{eqn:dw_update}
d\omega_t=\frac{\partial \tilde{\mathcal{M}}(f_{\omega_t},D_{dev})}{\partial \omega_t}=d\omega_{t+1}-\eta_t \frac{\partial^2L_{\mu_\theta(s_t)}(f_{\omega_t},D_{train}^t)}{\partial \omega_t^2} d\omega_{t+1}.
\end{equation}
At the same time, the gradient of $\tilde{\mathcal{M}}$ w.r.t. $\theta$ is accumulated at this time step as:
\begin{equation}
\label{eqn:theta_update}
d\theta =  d\theta -\eta_t \frac{\partial^2L_{\mu_\theta(s_t)}(f_{\omega_t}, D_{train}^t)}{\partial \theta \partial \omega_t}d\omega_{t+1}.
\end{equation}

We leave the detailed derivations for Eqn. (\ref{eqn:dw_update}) and (\ref{eqn:theta_update}) to Appendix. Furthermore it is worth-noting that the computing of $d\omega_t$ and $d\theta$ involves hessian vector product, which can be effectively computed via $\frac{\partial^2 g}{\partial x \partial y} v =\partial (\frac{\partial g}{\partial y} v)/\partial x$, without explicitly calculating the Hessian matrix. Reverting backwards from $t=T$ to $t=1$, we obtain $d\theta$ and then $\theta$ is updated using any gradient based optimization algorithm such as momentum SGD, forming one step optimization for $\theta$ which we call \emph{teacher optimization step}. By iterating teacher optimization steps we obtain the final teacher model. The details are listed in Algorithm~\ref{alg:train_teacher}.
\begin{algorithm}[ht]
\caption{Training Teacher Model $\mu_\theta$}
\label{alg:train_teacher}
\begin{algorithmic}
\State \textbf{Input}: Continuous relaxation $\tilde{m}$. Initial value of $\theta$.
\While{Teacher model parameter $\theta$ not converged} \Comment{One \emph{teacher optimization step}}
\State Randomly initialize student model parameter $\omega_0$.
\For {each time step $t=0,\cdots,T-1$} \Comment{Teach student model}
\State Conduct student model training step via Eqn. (\ref{eqn:sgd_seq}).
\EndFor 
\State $d\theta=0$. Compute $d\omega_T$ via Eqn. (\ref{eqn:dOmegaT}). 
\For {each time step $t=T-1,\cdots,0$}\Comment{Reversely calculating the gradient $d\theta$}
\State Update $d\theta$ as Eqn. (\ref{eqn:theta_update}).
\State Compute $d\omega_t$ as Eqn. (\ref{eqn:dw_update}).
\EndFor
\State Update $\theta$ using $d\theta$ via gradient based optimization algorithm.
\EndWhile
\State \textbf{Output}: the final teacher model $\mu_\theta$.
\end{algorithmic}
\end{algorithm}

\subsection{Discussion}

Another possible way to conduct teacher model optimization is through deep reinforcement learning. By treating the teacher model as a policy outputting continuous action (i.e., the loss function), one can leverage continuous control algorithm such as DDPG~\cite{ddpg} to optimize teacher model. However, reinforcement learning algorithms, including Q-learning based ones such as DDPG are sample inefficient, probably requiring huge amount of sampled trajectories to approximate the reward using a critic network. Considering the training of student model is typically costly, we resort to gradient based optimization algorithms instead. 

Furthermore, there are similarity between L2T-DLF and actor-critic (AC) method~\cite{ac1,ac2} in reinforcement learning (RL), in which a critic (corresponding to the parametric loss function) guides the optimization of an actor (corresponding to the student model). Apart from the difference within application domain (supervised learning versus RL), there are differences between the design principle of L2T-DLF and AC. For AC, by treating student model as actor, the student model output (e.g., $f_{\omega_t}(x_t)$) is essentially the action at timestep $t$, fed into the critic to output an approximation to the future reward (e.g., dev set accuracy). This is typically difficult since: 1) the student model output (i.e., the action) at a particular step $t$ is weakly related with the final dev performance. Therefore optimizing its action with the guidance from critic network is largely meaningless; 2) the approximation to the future reward is hard given the performance measure is highly non-smooth. As a comparison, L2T-DLF is more general in that at each timestep: 1) the teacher model considers the overall status of the student model for the sake of optimizing its parameters, rather than the instant action (i.e., the direct output); 2) the teacher model outputs a loss function with the goal of maximizing, but not approximating the future reward. In that sense, L2T-DLF is more appropriate to real world applications.



\section{Experiments}

We conduct comprehensive empirical verifications of the proposed L2T-DLF, in automatically discovering the most appropriate loss functions for student model training. The tasks in our experiments come from two domains: image classification, and neural machine translation.

\subsection{Image Classification}
\label{subsec:exp_image}

The evaluation measure $m$ here is the 0-1 accuracy: $m(y_1,y_2)=\mathbbm{1}_{y_1=y_2}$ where $\mathbbm{1}$ is the 0-1 indicator function.  The student model $f_\omega$ can be a logistic classifier specifying a softmax distribution $p_\omega(y|x)= \exp{(w_{y}'x +b_{y})}/\sum_{y^*\in\mathcal{Y}}{\exp{(w_{y^*}'x +b_{y^*})}}$ with $\omega=\{w_{y^*}, b_{y^*}\}_{y^*\in \mathcal{Y}}$. The class label is predicted as $\hat{y}=\arg\max_{y^*\in \mathcal{Y}}p_\omega(y^*|x)$ given input data $x$. Instead of imposing loss on $\hat{y}$ and ground-truth $y$, for the sake of efficient optimization $l$ typically takes the direct model output $p_\omega$ and $y$ as inputs. For example,  the most widely adopted loss function $l$ is cross-entropy loss $l(p_\omega, y)=-\log p_\omega(y|x)$, which could be re-written in vector form $l(p_\omega, y)=-\vec{y}'\log p_\omega$, where $\vec{y}\in\{0,1\}^{|\mathcal{Y}|}$ is a one-hot representation of the true label $y$, i.e., $\vec{y}_j=\mathbbm{1}_{j=y}, \forall j\in \mathcal{Y}$, $\vec{y}'$ is the transpose of $\vec{y}$ and $p_w\in \mathcal{R}^{|\mathcal{Y}|}$ is the probabilities for each class outputted via $f_\omega$. 

\begin{figure}
\centering
\subfigure[loss function]{
\begin{minipage}[b]{0.25\textwidth}
\includegraphics[width=1\linewidth]{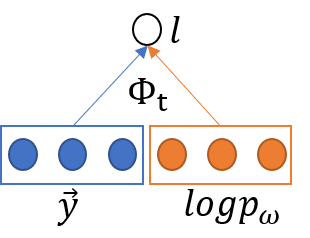}
\label{fig:student}
\end{minipage}
}
\subfigure[teacher model]{
\begin{minipage}[b]{0.60\textwidth}
\includegraphics[width=1\linewidth]{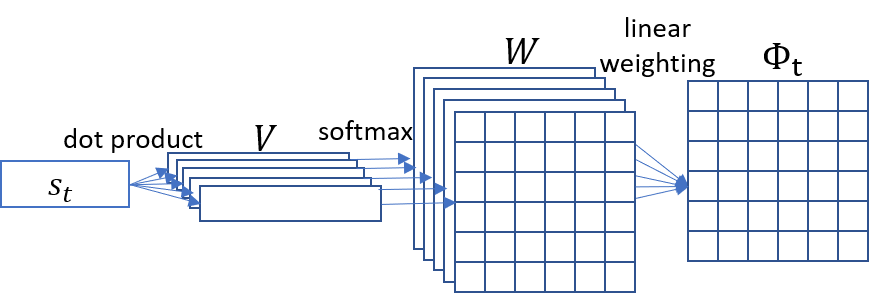}
\label{fig:teacher}
\end{minipage}
}
\caption{Left: the bilinear neural network specifying the loss function $l_{\Phi_t}(p_\omega,y)=-\sigma(\vec{y}'\Phi_t\log p_w)$. Right: the teacher model outputting $\Phi_t$ via attention mechanism:$\Phi_t=\mu_\theta(s_t)=Wsoftmax(Vs_t)$.}
\label{fig:teacher_student}
\end{figure}

Generalizing the cross entropy loss, we set the loss function coefficients $\Phi$ as a matrix interacting between $\log p_w$ and $\vec{y}$, which switches loss function at $t$-th timestep into $l_{\Phi_t}(p_\omega, y) = -\sigma(\vec{y}'\Phi_t\log p_w), \Phi_t \in \mathcal{R}^{|\mathcal{Y}|\times |\mathcal{Y}|}$, as is shown in Fig.~\ref{fig:student}. $\sigma$ is the sigmoid function. The teacher model $\mu_\theta$ here is then responsible for setting $\Phi_t$ according to the state feature vector of student model $s_t$: $\Phi_t=\mu_\theta(s_t)$. One possible form of the teacher model is a neural network with attention mechanism (shown in Fig.~\ref{fig:teacher}): $\Phi_t=\mu_\theta(s_t)=Wsoftmax(Vs_t)$, where $W\in \mathcal{R}^{|\mathcal{Y}|\times |\mathcal{Y}|\times N}, V\in \mathcal{R}^{N\times |s_t|}$ constitute the teacher model parameter set $\theta$, $N=10$ is the number of keys in attention mechanism. The state vector $s_t$ is a $13$ dimensional vector composing of 1) the current iteration number $t$; 2) current training accuracy of $f_\omega$; 3) current dev accuracy of $f_\omega$; 4) current precision of $f_\omega$ for the $10$ classes on the dev set, all normalized into $[0,1]$.


We choose three widely adopted datasets: the MNIST, CIFAR-10 and CIFAR-100 datasets. For the sake of showing the robustness of L2T-DLF, the student models we choose cover a wide range, including multi-layer perceptron (MLP), plain convolutional neural network (CNN) following LeNet architecture~\cite{LeNet}, and advanced CNN architecture including ResNet~\cite{ResNet}, Wide-ResNet~\cite{WRN} and DenseNet~\cite{DenseNet}. For all the student models, we use momentum stochastic gradient descent to perform training. In Appendix we describe the network structures of student models. 


The different loss functions we compare include: 1) Cross entropy loss $L_{ce}(p_\omega(x), y)=-\log p_\omega(y|x)$, which is the most widely adopted loss function to train neural network model; 2) The smooth 0-1 loss proposed in~\cite{smooth_acc}. It optimizes a smooth version of 0-1 accuracy in binary classification. We extend it to handle multi-class case by modifying the loss function as $L_{smooth}(p_\omega(x), y)=-\log \sigma(K(\log p_\omega(y|x)-\max_{y^*\neq y}\log p_\omega(y^*|x)))$. It is not difficult to observe when $K\rightarrow +\infty$,  $-L_{smooth}$ exactly matches the 0-1 accuracy. We choose the value of $K$ to be $50$ according to the performance on dev set; 3) The large-margin softmax loss in~\cite{lm_softmax} denoted as $L_{lm}$, which aims to enhance discrimination between different classes via maximizing the margin induced by the angle between $x$ and a target class representation $w_y$. We use the open-sourced code released by the authors in our experiment; 4) The loss function discovered via the teacher in L2T-DLF. The teacher models are optimized with Adam~\cite{adam} and the detailed setting is in Appendix.


\begin{table}[]
\centering
\caption{The recognition results (error rate \%) on MNIST dataset.}
\label{tbl:MNIST}
\begin{tabular}{cccccc}
\toprule
\begin{tabular}[c]{@{}c@{}}Student Model/\\  Loss\end{tabular} & \begin{tabular}[c]{@{}c@{}}Cross\\ Entropy~\cite{cross_entropy} \end{tabular} & Smooth~\cite{smooth_acc} & \begin{tabular}[c]{@{}c@{}}Large-Margin \\ Softmax~\cite{lm_softmax} \end{tabular} & \begin{tabular}[c]{@{}c@{}}L2T-DLF \end{tabular} \\ 
\midrule
\emph{MLP}  &  1.94   & 1.89   &  1.83  &  \textbf{1.69}   \\
\emph{LeNet}  &  0.98  &  0.94  &  0.88   &   \textbf{0.77}   \\  
\bottomrule
\end{tabular}
\end{table}

\begin{table}[]
\centering
\caption{The recognition results (error rate \%) on CIFAR-10 (C10) and CIFAR-100 (C100) dataset}
\label{tbl:C10-100}
\begin{tabular}{l c c c c}
\toprule
\begin{tabular}[c]{@{}c@{}}Student Model/\\  Loss\end{tabular} & \begin{tabular}[c]{@{}c@{}}Cross\\ Entropy~\cite{cross_entropy} \end{tabular} & Smooth~\cite{smooth_acc} & \begin{tabular}[c]{@{}c@{}}Large-Margin \\ Softmax~\cite{lm_softmax} \end{tabular} &  \begin{tabular}[c]{@{}c@{}}L2T-DLF \end{tabular} \\ 
\midrule
  & C10/C100 & C10/C100 & C10/C100 & C10/C100 \\
\cmidrule{2-5}  
\emph{ResNet-8}  & 12.45/39.79 & 12.08/39.52 & 11.34/38.93 & \textbf{10.82/38.27} \\ 
\emph{ResNet-20}  & 8.75/32.33 & 8.53/32.01 & 8.02/31.65 & \textbf{7.63/30.97}  \\ 
\emph{ResNet-32}  & 7.51/30.38  & 7.42/30.12 & 7.01/29.56 & \textbf{6.95/29.25} \\ 
\emph{WRN}  & 3.80/- & 3.81/- & 3.69/- & \textbf{3.42}/- \\ 
\emph{DenseNet-BC}  & 3.54/- & 3.48/- & 3.37/- & \textbf{3.08}/- \\ 
\bottomrule
\end{tabular}
\end{table}



The classification results on MNIST, CIFAR-10 and CIFAR-100 are respectively shown in Table \ref{tbl:MNIST} and~\ref{tbl:C10-100}. As can be observed, on all the three tasks, the dynamic loss functions outputted via teacher model help to cultivate better student model. For example, the teacher model helps WRN to achieve $3.42\%$ classification error rate on CIFAR-10, which is on par with the result discovered via automatic architecture search (e.g., $3.41\%$ of NASNet~\cite{NAS}). Furthermore, our dynamic loss functions for DenseNet on CIFAR-10 reduces the error rate of DenseNet-BC ($k$=$40$) from $3.54\%$ to $3.08\%$, where the gain is a non-trival margin.  

\subsubsection{Teacher Optimization}

\begin{figure}[]
\centering
  \includegraphics[width=0.55\linewidth]{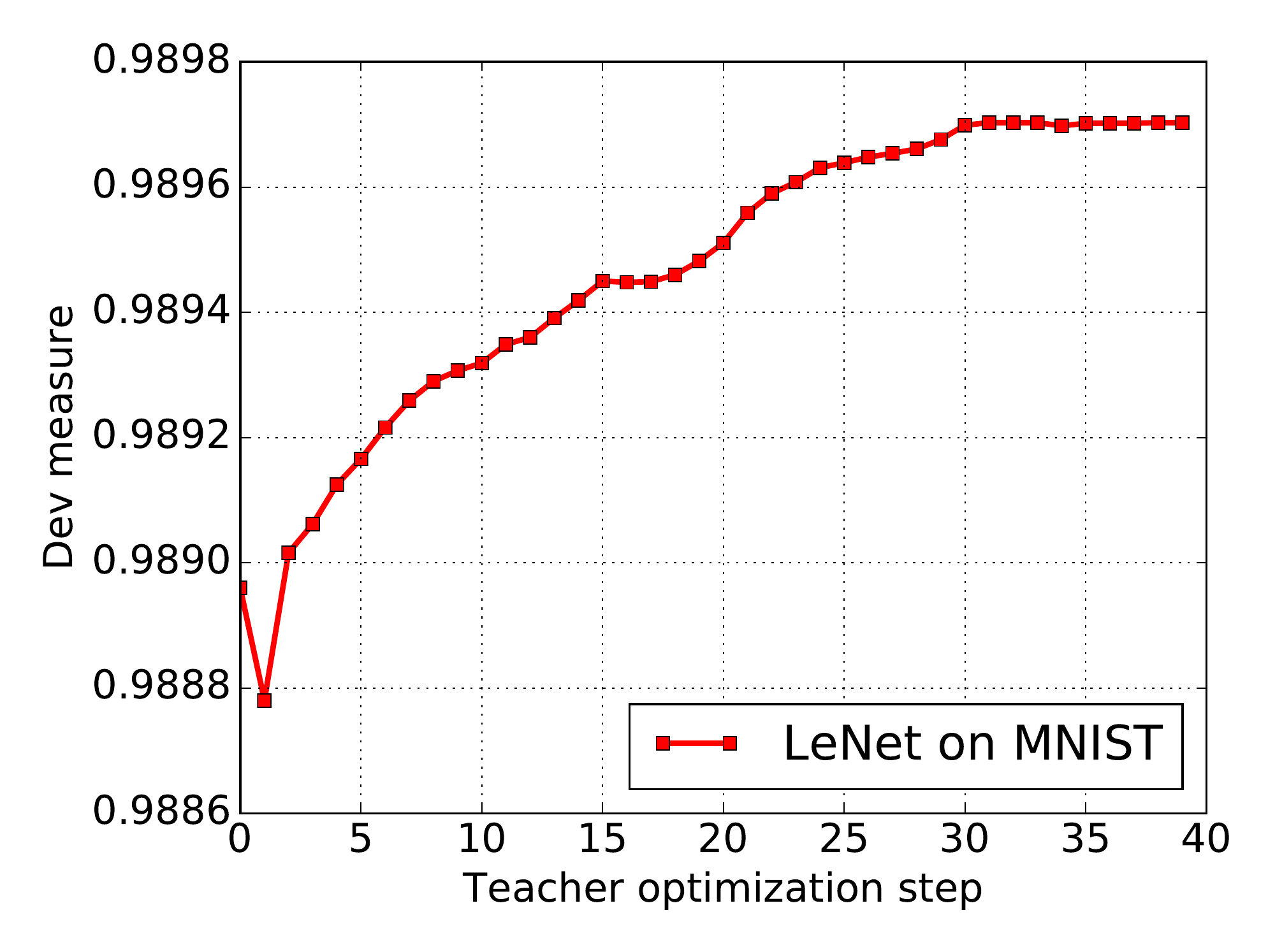}
  \caption{Measure score on the MNIST dev set along the teacher model optimization. The student model is LeNet.}
  \label{fig:mnist_measure}
\end{figure}

In Fig. \ref{fig:mnist_measure}, we provide the dev measure performance along with the teacher model optimization in MNIST experiment, the student model is LeNet. It can be observed that the dev measure is increasing along with the teacher model optimizing, and finally converges to a high score. 

\subsubsection{Analysis Towards the Loss Functions}

\begin{figure}
\centering
\subfigure[Class \emph{Otter}]{
\begin{minipage}[b]{0.48\textwidth}
\includegraphics[width=1\linewidth]{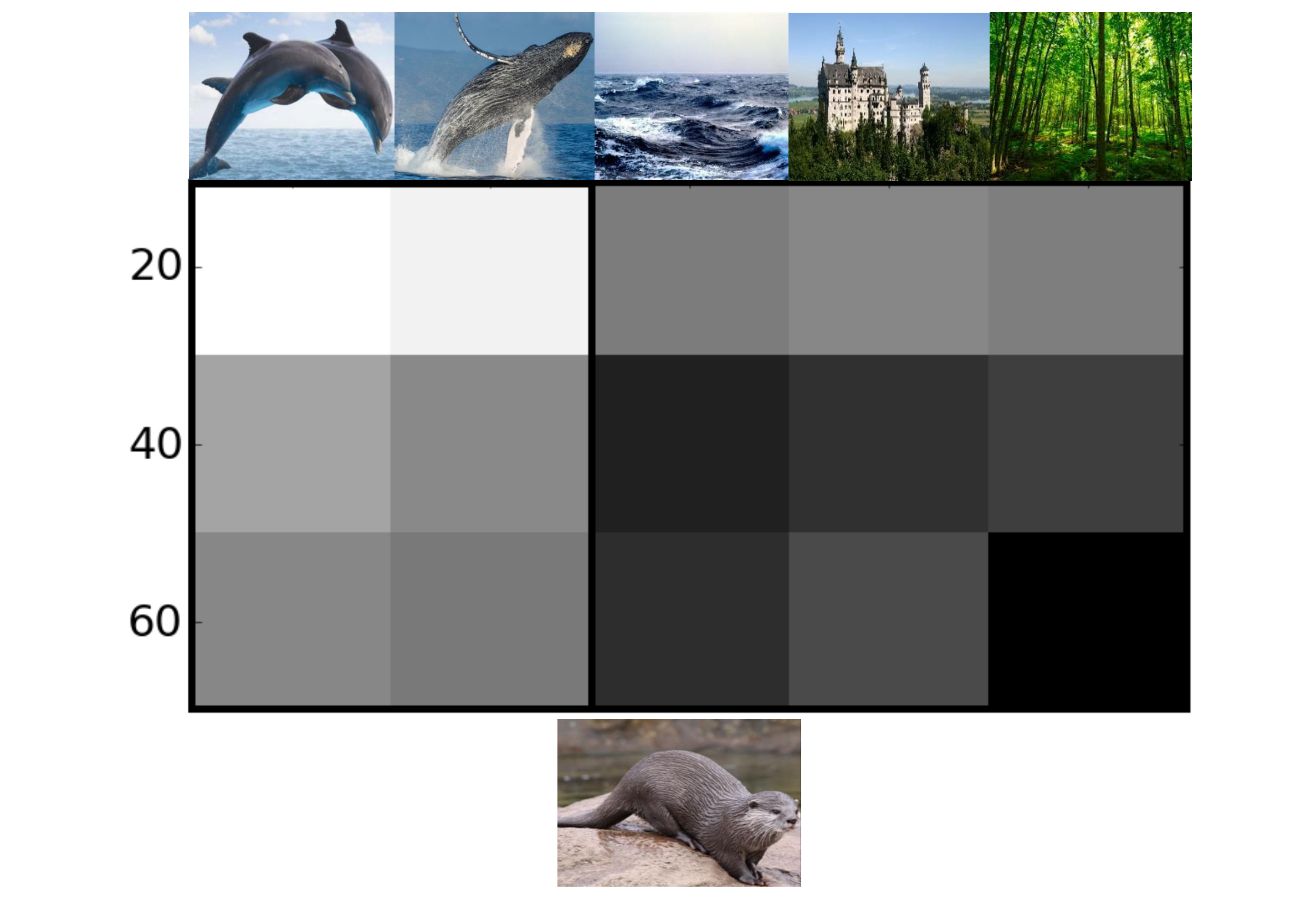}
\label{fig:otter}
\end{minipage}
}
\subfigure[Class \emph{Baby}]{
\begin{minipage}[b]{0.48\textwidth}
\includegraphics[width=1\linewidth]{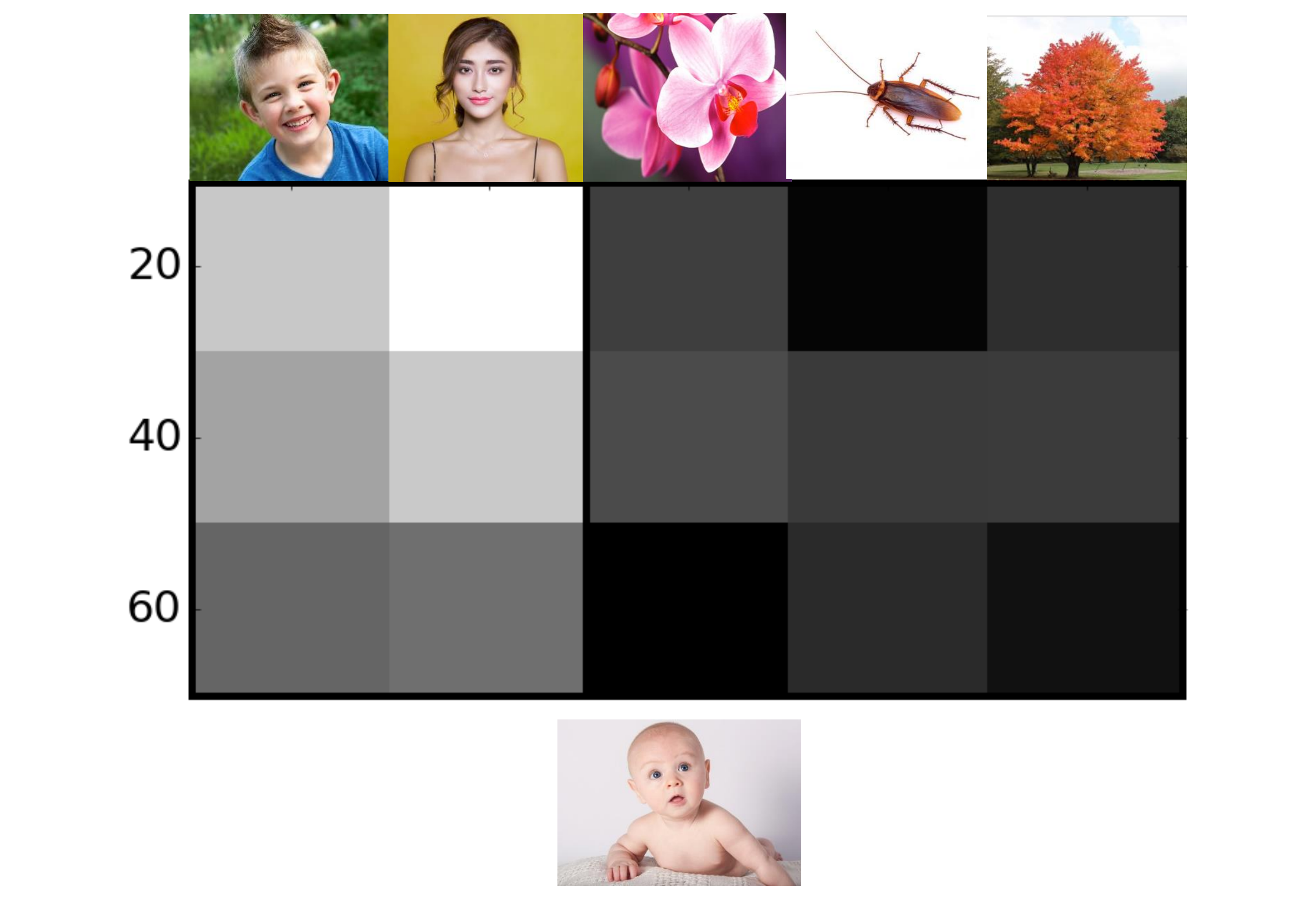}
\label{fig:baby}
\end{minipage}
}
\caption{Coefficient matrix $\Phi_t$ outputted via teacher model. The y-axis ($20, 40, 60$) corresponds to the different epochs of the student model training. Darker color means the coefficients value are more negative while shallower color means more positive. In each figure, the leftmost two columns denote similar classes and the rightmost three columns represent dissimilar classes.}
\label{fig:ana_lf}
\end{figure}

To better understand the loss functions outputted via teacher model, we visualize the coefficients of some loss functions outputted by teacher model for training ResNet-8 in CIFAR-100 classification task. Specifically, note that the loss function $l_{\Phi_t}(p_\omega, y) = -\sigma(\vec{y}'\Phi_t\log p_w)$ essentially characterizes the \emph{correlations among different classes} via the coefficients $\Phi_t$. Positive $\Phi_t(i,j)$ value means positive correlation between class $i$ and $j$ that their probabilities should be jointly maximized whereas negative value imposes negative correlation and higher discrimination between the two classes $i$ and $j$. We choose two classes in CIFAR-100: the \emph{Otter} and \emph{Baby} as class $i$ and for each of them pick several representative classes as class $j$. The corresponding $\Phi_t(i,j)$ values are visualized in Fig.~\ref{fig:ana_lf}, with $t=20,40,60$ denoting the coefficients outputted via teacher model at $t$-th epoch of student model training. As can be observed, at the initial phase of training student model ($t=20$), the teacher model chooses to enhance the correlation between two similar classes, e.g, \emph{Otter} and \emph{Dolphin}, \emph{Baby} and \emph{Boy}, for the sake of speeding up training. Comparatively, when the student model is powerful enough ($t=60$), the teacher model will force it to perform better in discriminating two similar classes, as indicated via the more negative coefficient values $\Phi_t(i,j)$. The variation of $\Phi_t(i,j)$ values w.r.t. $t$ well demonstrates the teacher model captures the status of student model in outputting correspondingly appropriate loss functions.

\subsection{Neural Machine Translation}
\label{subsec:nmt}

In the task of \emph{neural machine translation} (NMT), the evaluation measure $m(\hat{y}, y)$ is typically the BLEU score~\cite{BLEU} between the translated sentence $\hat{y}$ and ground-truth reference $y$. The student model $f_\omega$ is a neural network performing sequence-to-sequence generation based on models including RNN~\cite{s2s}, CNN~\cite{conv_s2s} and self-attention network~\cite{transformer}. The decoding process of $f_\omega$ is typically \emph{autoregressive}, in that $f_\omega$ factorizes the translation probability as $p_\omega(y|x)=\prod_{r=1}^{|y|}p_\omega (y_r|x,y_{<r})$. Here $p_\omega(\cdot|x, y_{<r})$ is the distribution on target vocabulary $\mathcal{V}$ at the $r$-th position, taking the source side sentence $x$ and the previous words $y_{<r}$ as inputs. Similar to the classification task, the loss function generalizing cross entropy loss is $l_\Phi=-\sum_{r=1}^{|y|}\sigma(\vec{y}_r' diag(\Phi) \log p_\omega(\cdot|x,y_{<r}))$, where $\Phi\in\mathcal{R}^{|\mathcal{V}|}$ is the coefficients of the loss function and $diag(\Phi)$ denotes the diagnoal matrix with $\Phi$ as its diagonal elements. Here we set the interaction matrix as diagonal mainly for the sake of computational efficiency, since the target vocabulary size $|\mathcal{V}|$ is usally very large (e.g., $30k$). The teacher model then outputs $\Phi_t$ at timestep $t$ taking $s_t$ as input: $\Phi_t=\mu_\theta(s_t)=Wsoftmax(Vs_t)$, where teacher model parameter $\theta= \{W\in \mathcal{R}^{|\mathcal{V}|\times N}, V\in \mathcal{R}^{N\times |s_t|}\}$. We set $N=5$ and for the state vector $s_t$, it is the same with that in classification except: 1) the training/dev set accuracy is now replaced with BLEU scores; 2) the last ten features in $s_t$ for classification are ignored, leading to $|s_t|=3$.

We choose a widely used benchmark dataset in NMT literature \cite{sequence_level_rnn,wu2017adversarial,wu2018study}, released in IWSLT-14 German-English evaluation campaign~\cite{IWSLT14}, as the test-bed for different loss functions. The student model $f_\omega$ for this task is based on LSTM with attention~\cite{s2s_attn}. For the sake of fair comparison with previous works~\cite{ac_s2s,sequence_level_rnn}, we use single layer LSTM model as $f_\omega$ and name it as \emph{LSTM-1}. To further verify the effectiveness of L2T-DLF, we use a deeper translation model stacking two LSTM layers as $f_\omega$. We denote such stronger student model as \emph{LSTM-2}. 
Furthermore, we also evaluate our L2T-DLF on the Transformer~\cite{transformer} network. The Transformer architecture is based on the self-attention mechanism \cite{lin2017structured}, and it achieves superior performance on several NMT tasks.
Both LSTM/Transformer student models are trained with simple SGD. In Appendix we provide the details of the LSTM/Transformer student models and the training settings of student/teacher models.
 
\label{subsec:exp_nmt}
\begin{table}[]
\centering
\caption{The translation results (BLEU score) on IWSLT-14 German-English task.}
\label{tbl:nmt}
\begin{tabular}{cccccc}
\toprule
\begin{tabular}[c]{@{}c@{}}Student Model/\\  Loss\end{tabular} & \begin{tabular}[c]{@{}c@{}}Cross\\ Entropy~\cite{teacher_forcing} \end{tabular} & RL~\cite{sequence_level_rnn} & AC~\cite{ac_s2s} & Softmax-Margin~\cite{classical_loss_nmt} & \begin{tabular}[c]{@{}c@{}}L2T-DLF \end{tabular} \\  
\midrule
\emph{LSTM-1}  &  27.28  &  27.53  & 27.75  &  28.12  &  \textbf{29.52}  \\ 
\emph{LSTM-2}  &  30.86  &  31.03  &  31.21  &  31.22  & \textbf{31.75}  \\ 
\emph{Transformer} & 34.01 & 34.32 &  34.34 & 34.46 & \textbf{34.80} \\
\bottomrule
\end{tabular}
\end{table}

The loss functions we leverage to train student models include: 1) Cross entropy loss $L_{ce}$ to perform maximum likelihood estimation (MLE) for training LSTM/Transformer model with teacher forcing~\cite{teacher_forcing}; 2) The reinforcement learning (RL) loss $L_{rl}$, a.k.a, sequence level training~\cite{sequence_level_rnn} or minimum risk training~\cite{rl4nmt}, targets at directly optimizing the BLEU scores for NMT models. A typical RL loss is $L_{rl}(p_\omega(x), y)=-\sum_{y^*\in \mathcal{Y}}\log p_\omega(y^*|x)(BLEU(y^*,y) - b)$, where $b$ is the reward baseline and $\mathcal{Y}$ is the candidate subset; 3) The loss specified via actor-critic (AC) algorithm $L_{ac}$~\cite{ac_s2s}, which approximates the BLEU score via a critic network; 4) The softmax-margin loss, which is empirically shown to be the most effective structural prediction loss for NMT~\cite{classical_loss_nmt}; 5) The loss function discovered via our L2T-DLF. 

We report the experimental results in Table~\ref{tbl:nmt}. From the table, we can clearly observe the dynamic loss functions outputted via our teacher model can guide the student model to have superior performance compared with other specially designed loss functions. Specifically, with a shallow student model \emph{LSTM-1}, we improve the BLEU score by more than $2.0$ points compared with predefined cross-entropy loss. In addition, our \emph{LSTM-2} student model achieves $31.75$ BLEU score and it surpasses previously reported best result $30.08$ by \cite{NPBMT} on IWSLT-14 German-English achieved via RNN/LSTM models. With a much stronger \emph{Transformer} student model, we also improve the model performance from BLEU score $34.01$ to $34.80$. The above results clearly demonstrate the effectiveness of our L2T-DLF approach.

\section{Conclusion}

In contrast to expert designed and fixed loss functions in conventional machine learning systems, we in this paper study how to learn dynamic loss functions so as to better teach a student machine learning model.  Since loss functions provided by the teacher model dynamically change with respect to the growth of the student model and the teacher model is trained through end-to-end optimization, the quality of the student model gets improved significantly, as shown in our experiments. We hope our work will stimulate and inspire the research community to automatically discover loss functions better than expert designed ones. As to future work, we would like to conduct empirical verification on tasks with more powerful student models and larger datasets. We are also interested in trying more complicated teacher models such as deeper neural networks. 

\section{Acknowledgement}
This work was partially supported by the NSFC 61573387. We thank all the anonymous reviewers for their constructive feedbacks.

\bibliography{l2t_loss}

\begin{thebibliography}{10}

\bibitem{teach_old_1}
John~R Anderson, C~Franklin Boyle, and Brian~J Reiser.
\newblock Intelligent tutoring systems.
\newblock {\em Science}, 228(4698):456--462, 1985.

\bibitem{learning2learn_sgd}
Marcin Andrychowicz, Misha Denil, Sergio Gomez, Matthew~W Hoffman, David Pfau,
  Tom Schaul, and Nando de~Freitas.
\newblock Learning to learn by gradient descent by gradient descent.
\newblock In {\em Advances in Neural Information Processing Systems}, pages
  3981--3989, 2016.

\bibitem{ac_s2s}
Dzmitry Bahdanau, Philemon Brakel, Kelvin Xu, Anirudh Goyal, Ryan Lowe, Joelle
  Pineau, Aaron Courville, and Yoshua Bengio.
\newblock An actor-critic algorithm for sequence prediction.
\newblock {\em arXiv preprint arXiv:1607.07086}, 2016.

\bibitem{s2s_attn}
Dzmitry Bahdanau, Kyunghyun Cho, and Yoshua Bengio.
\newblock Neural machine translation by jointly learning to align and
  translate.
\newblock {\em arXiv preprint arXiv:1409.0473}, 2014.

\bibitem{ac1}
Andrew~G Barto, Richard~S Sutton, and Charles~W Anderson.
\newblock Neuronlike adaptive elements that can solve difficult learning
  control problems.
\newblock {\em IEEE transactions on systems, man, and cybernetics},
  (5):834--846, 1983.

\bibitem{autoDiff}
Atilim~Gunes Baydin and Barak~A Pearlmutter.
\newblock Automatic differentiation of algorithms for machine learning.
\newblock {\em arXiv preprint arXiv:1404.7456}, 2014.

\bibitem{RMD}
Yoshua Bengio.
\newblock Gradient-based optimization of hyperparameters.
\newblock {\em Neural computation}, 12(8):1889--1900, 2000.

\bibitem{CL}
Yoshua Bengio, J{\'e}r{\^o}me Louradour, Ronan Collobert, and Jason Weston.
\newblock Curriculum learning.
\newblock In {\em Proceedings of the 26th ICML}, pages 41--48. ACM, 2009.

\bibitem{IWSLT14}
M~Cettolo, J~Niehues, S~St{\"u}ker, L~Bentivogli, and M~Federico.
\newblock Report on the 11th iwslt evaluation campaign, iwslt 2014.
\newblock In {\em IWSLT-International Workshop on Spoken Language Processing},
  pages 2--17. Marcello Federico, Sebastian St{\"u}ker, Fran{\c{c}}ois Yvon,
  2014.

\bibitem{l2l_wo_gd}
Yutian Chen, Matthew~W Hoffman, Sergio~G{\'o}mez Colmenarejo, Misha Denil,
  Timothy~P Lillicrap, Matt Botvinick, and Nando de~Freitas.
\newblock Learning to learn without gradient descent by gradient descent.
\newblock {\em arXiv preprint arXiv:1611.03824}, 2016.

\bibitem{cross_entropy}
Pieter-Tjerk De~Boer, Dirk~P Kroese, Shie Mannor, and Reuven~Y Rubinstein.
\newblock A tutorial on the cross-entropy method.
\newblock {\em Annals of operations research}, 134(1):19--67, 2005.

\bibitem{classical_loss_nmt}
Sergey Edunov, Myle Ott, Michael Auli, David Grangier, and Marc'Aurelio
  Ranzato.
\newblock Classical structured prediction losses for sequence to sequence
  learning.
\newblock {\em arXiv preprint arXiv:1711.04956}, 2017.

\bibitem{l2t_iclr}
Yang Fan, Fei Tian, Tao Qin, Xiang-Yang Li, and Tie-Yan Liu.
\newblock Learning to teach.
\newblock In {\em International Conference on Learning Representations}, 2018.

\bibitem{MAML}
Chelsea Finn, Pieter Abbeel, and Sergey Levine.
\newblock Model-agnostic meta-learning for fast adaptation of deep networks.
\newblock In Doina Precup and Yee~Whye Teh, editors, {\em Proceedings of the
  34th International Conference on Machine Learning}, volume~70 of {\em
  Proceedings of Machine Learning Research}, pages 1126--1135, International
  Convention Centre, Sydney, Australia, 06--11 Aug 2017. PMLR.

\bibitem{forwardRMD}
Luca Franceschi, Michele Donini, Paolo Frasconi, and Massimiliano Pontil.
\newblock Forward and reverse gradient-based hyperparameter optimization.
\newblock In Doina Precup and Yee~Whye Teh, editors, {\em Proceedings of the
  34th International Conference on Machine Learning}, volume~70 of {\em
  Proceedings of Machine Learning Research}, pages 1165--1173, International
  Convention Centre, Sydney, Australia, 06--11 Aug 2017. PMLR.

\bibitem{conv_s2s}
Jonas Gehring, Michael Auli, David Grangier, Denis Yarats, and Yann~N Dauphin.
\newblock Convolutional sequence to sequence learning.
\newblock In {\em International Conference on Machine Learning}, pages
  1243--1252, 2017.

\bibitem{teach_old_2}
S.A. Goldman and M.J. Kearns.
\newblock On the complexity of teaching.
\newblock {\em J. Comput. Syst. Sci.}, 50(1):20--31, February 1995.

\bibitem{easy_l2t_0}
Chen Gong, Dacheng Tao, Wei Liu, Liu Liu, and Jie Yang.
\newblock Label propagation via teaching-to-learn and learning-to-teach.
\newblock {\em IEEE transactions on neural networks and learning systems},
  28(6):1452--1465, 2017.

\bibitem{easy_l2t_1}
Chen Gong, Dacheng Tao, Jie Yang, and Wei Liu.
\newblock Teaching-to-learn and learning-to-teach for multi-label propagation.
\newblock In {\em AAAI 2016}, pages 1610--1616, 2016.

\bibitem{grad_opt}
Elad Hazan, Kfir~Yehuda Levy, and Shai Shalev-Shwartz.
\newblock On graduated optimization for stochastic non-convex problems.
\newblock In {\em International Conference on Machine Learning}, pages
  1833--1841, 2016.

\bibitem{ResNet}
Kaiming He, Xiangyu Zhang, Shaoqing Ren, and Jian Sun.
\newblock Deep residual learning for image recognition.
\newblock In {\em Proceedings of the IEEE conference on computer vision and
  pattern recognition}, pages 770--778, 2016.

\bibitem{henderson2016end}
Paul Henderson and Vittorio Ferrari.
\newblock End-to-end training of object class detectors for mean average
  precision.
\newblock In {\em Asian Conference on Computer Vision}, pages 198--213.
  Springer, 2016.

\bibitem{ShowingDemo}
Mark~K Ho, Michael Littman, James MacGlashan, Fiery Cushman, and Joseph~L
  Austerweil.
\newblock Showing versus doing: Teaching by demonstration.
\newblock In {\em Advances in Neural Information Processing Systems 29}, pages
  3027--3035. 2016.

\bibitem{lstm}
Sepp Hochreiter and J{\"u}rgen Schmidhuber.
\newblock Long short-term memory.
\newblock {\em Neural computation}, 9(8):1735--1780, 1997.

\bibitem{DenseNet}
Gao Huang, Zhuang Liu, Laurens Van Der~Maaten, and Kilian~Q Weinberger.
\newblock Densely connected convolutional networks.
\newblock In {\em CVPR}, 2017.

\bibitem{NPBMT}
Po-Sen Huang, Chong Wang, Sitao Huang, Dengyong Zhou, and Li~Deng.
\newblock Towards neural phrase-based machine translation.
\newblock In {\em International Conference on Learning Representations}, 2018.

\bibitem{adam}
Diederik~P Kingma and Jimmy Ba.
\newblock Adam: A method for stochastic optimization.
\newblock {\em arXiv preprint arXiv:1412.6980}, 2014.

\bibitem{SPL}
M~Pawan Kumar, Benjamin Packer, and Daphne Koller.
\newblock Self-paced learning for latent variable models.
\newblock In {\em Advances in Neural Information Processing Systems}, pages
  1189--1197, 2010.

\bibitem{LeNet}
Yann LeCun, L{\'e}on Bottou, Yoshua Bengio, and Patrick Haffner.
\newblock Gradient-based learning applied to document recognition.
\newblock {\em Proceedings of the IEEE}, 86(11):2278--2324, 1998.

\bibitem{learning_to_optimize}
Ke~Li and Jitendra Malik.
\newblock Learning to optimize.
\newblock {\em arXiv preprint arXiv:1606.01885}, 2016.

\bibitem{uneven_margin}
Yaoyong Li, Hugo Zaragoza, Ralf Herbrich, John Shawe-Taylor, and Jaz~S.
  Kandola.
\newblock The perceptron algorithm with uneven margins.
\newblock In {\em Proceedings of the Nineteenth International Conference on
  Machine Learning}, ICML '02, pages 379--386, San Francisco, CA, USA, 2002.
  Morgan Kaufmann Publishers Inc.

\bibitem{ddpg}
Timothy~P Lillicrap, Jonathan~J Hunt, Alexander Pritzel, Nicolas Heess, Tom
  Erez, Yuval Tassa, David Silver, and Daan Wierstra.
\newblock Continuous control with deep reinforcement learning.
\newblock {\em arXiv preprint arXiv:1509.02971}, 2015.

\bibitem{FocalLoss}
Tsung-Yi Lin, Priya Goyal, Ross Girshick, Kaiming He, and Piotr Doll{\'a}r.
\newblock Focal loss for dense object detection.
\newblock {\em arXiv preprint arXiv:1708.02002}, 2017.

\bibitem{lin2017structured}
Zhouhan Lin, Minwei Feng, Cicero Nogueira~dos Santos, Mo~Yu, Bing Xiang, Bowen
  Zhou, and Yoshua Bengio.
\newblock A structured self-attentive sentence embedding.
\newblock In {\em ICLR}, 2017.

\bibitem{teaching+dimension}
Ji~Liu and Xiaojin Zhu.
\newblock The teaching dimension of linear learners.
\newblock {\em Journal of Machine Learning Research}, 17(162):1--25, 2016.

\bibitem{iterative_teaching}
Weiyang Liu, Bo~Dai, James Rehg, and Le~Song.
\newblock Iterative machine teaching.
\newblock In {\em Proceedings of the 34st International Conference on Machine
  Learning (ICML-17)}, pages 1188--1196, 2017.

\bibitem{lm_softmax}
Weiyang Liu, Yandong Wen, Zhiding Yu, and Meng Yang.
\newblock Large-margin softmax loss for convolutional neural networks.
\newblock In {\em International Conference on Machine Learning}, pages
  507--516, 2016.

\bibitem{NAO}
Renqian Luo, Fei Tian, Tao Qin, and Tie-Yan Liu.
\newblock Neural architecture optimization.
\newblock {\em arXiv preprint arXiv:1808.07233}, 2018.

\bibitem{RMD_rev}
Dougal Maclaurin, David Duvenaud, and Ryan Adams.
\newblock Gradient-based hyperparameter optimization through reversible
  learning.
\newblock In {\em International Conference on Machine Learning}, pages
  2113--2122, 2015.

\bibitem{openAIpeda}
Smitha Milli, Pieter Abbeel, and Igor Mordatch.
\newblock Interpretable and pedagogical examples.
\newblock {\em arXiv preprint arXiv:1711.00694}, 2017.

\bibitem{smooth_acc}
Tan Nguyen and Scott Sanner.
\newblock Algorithms for direct 0--1 loss optimization in binary
  classification.
\newblock In {\em International Conference on Machine Learning}, pages
  1085--1093, 2013.

\bibitem{BLEU}
Kishore Papineni, Salim Roukos, Todd Ward, and Wei-Jing Zhu.
\newblock Bleu: a method for automatic evaluation of machine translation.
\newblock In {\em Proceedings of the 40th annual meeting on association for
  computational linguistics}, pages 311--318. Association for Computational
  Linguistics, 2002.

\bibitem{momentum}
Ning Qian.
\newblock On the momentum term in gradient descent learning algorithms.
\newblock {\em Neural networks}, 12(1):145--151, 1999.

\bibitem{sequence_level_rnn}
Marc'Aurelio Ranzato, Sumit Chopra, Michael Auli, and Wojciech Zaremba.
\newblock Sequence level training with recurrent neural networks.
\newblock {\em arXiv preprint arXiv:1511.06732}, 2015.

\bibitem{metaLearning}
Jurgen Schmidhuber.
\newblock Evolutionary principles in self-referential learning.
\newblock {\em Diploma thesis, Institut f. Informatik, Tech. Univ. Munich},
  1987.

\bibitem{BPE}
Rico Sennrich, Barry Haddow, and Alexandra Birch.
\newblock Neural machine translation of rare words with subword units.
\newblock In {\em Proceedings of the 54th Annual Meeting of the Association for
  Computational Linguistics (Volume 1: Long Papers)}, volume~1, pages
  1715--1725, 2016.

\bibitem{Pedagogical}
Patrick Shafto, Noah~D Goodman, and Thomas~L Griffiths.
\newblock A rational account of pedagogical reasoning: Teaching by, and
  learning from, examples.
\newblock {\em Cognitive psychology}, 71:55--89, 2014.

\bibitem{rl4nmt}
Shiqi Shen, Yong Cheng, Zhongjun He, Wei He, Hua Wu, Maosong Sun, and Yang Liu.
\newblock Minimum risk training for neural machine translation.
\newblock In {\em Proceedings of the 54th Annual Meeting of the Association for
  Computational Linguistics (Volume 1: Long Papers)}, pages 1683--1692.
  Association for Computational Linguistics, 2016.

\bibitem{direct_opt}
Yang Song, Alexander Schwing, Raquel Urtasun, et~al.
\newblock Training deep neural networks via direct loss minimization.
\newblock In {\em International Conference on Machine Learning}, pages
  2169--2177, 2016.

\bibitem{s2s}
Ilya Sutskever, Oriol Vinyals, and Quoc~V Le.
\newblock Sequence to sequence learning with neural networks.
\newblock In {\em Advances in neural information processing systems}, pages
  3104--3112, 2014.

\bibitem{ac2}
Richard~Stuart Sutton.
\newblock Temporal credit assignment in reinforcement learning.
\newblock 1984.

\bibitem{softrank}
Michael Taylor, John Guiver, Stephen Robertson, and Tom Minka.
\newblock Softrank: optimizing non-smooth rank metrics.
\newblock In {\em Proceedings of the 2008 International Conference on Web
  Search and Data Mining}, pages 77--86. ACM, 2008.

\bibitem{L2LBook}
Sebastian Thrun and Lorien Pratt.
\newblock {\em Learning to learn}.
\newblock Springer Science \& Business Media, 2012.

\bibitem{transformer}
Ashish Vaswani, Noam Shazeer, Niki Parmar, Jakob Uszkoreit, Llion Jones,
  Aidan~N Gomez, {\L}ukasz Kaiser, and Illia Polosukhin.
\newblock Attention is all you need.
\newblock In {\em Advances in Neural Information Processing Systems}, pages
  6000--6010, 2017.

\bibitem{teacher_forcing}
Ronald~J Williams and David Zipser.
\newblock A learning algorithm for continually running fully recurrent neural
  networks.
\newblock {\em Neural computation}, 1(2):270--280, 1989.

\bibitem{wu2018study}
Lijun Wu, Fei Tian, Tao Qin, Jianhuang Lai, and Tie-Yan Liu.
\newblock A study of reinforcement learning for neural machine translation.
\newblock In {\em EMNLP}, 2018.

\bibitem{wu2017adversarial}
Lijun Wu, Yingce Xia, Li~Zhao, Fei Tian, Tao Qin, Jianhuang Lai, and Tie-Yan
  Liu.
\newblock Adversarial neural machine translation.
\newblock In {\em ACML}, 2018.

\bibitem{WRN}
Sergey Zagoruyko and Nikos Komodakis.
\newblock Wide residual networks.
\newblock {\em arXiv preprint arXiv:1605.07146}, 2016.

\bibitem{machine_teaching}
Xiaojin Zhu.
\newblock Machine teaching: An inverse problem to machine learning and an
  approach toward optimal education.
\newblock In {\em Proceedings of the Twenty-Ninth AAAI Conference on Artificial
  Intelligence}, AAAI'15, pages 4083--4087. AAAI Press, 2015.

\bibitem{NAS}
Barret Zoph and Quoc Le.
\newblock Neural architecture search with reinforcement learning.
\newblock In {\em International Conference on Learning Representations}, 2017.

\end{thebibliography}

\newpage
\appendix

\section{Derivations For the Updating Rules of Teacher Model Parameters}
We provide derivations of Eqn. (4) and (5) in the original paper. The starting point is Eqn.(1):

\begin{equation}
\label{eqn:sgd_seq}
\omega_{t+1} = \omega_t - \eta_t \frac{\partial L_{\Phi_t}(f_{w_t}, D_{train}^t)}{\partial \omega_t}=\omega_t-\eta_t\frac{\partial L_{\mu_{\theta}(s_t)}(f_{w_t}, D_{train}^t)}{\partial \omega_t}.
\end{equation}

Then we have:

\begin{equation}
\label{eqn:dw_update}
\begin{aligned}
d\omega_t=\frac{\partial \tilde{\mathcal{M}}(f_{\omega_T}, D_{dev})}{\partial \omega_t}=&(\frac{\partial \omega_{t+1}}{\partial \omega_{t}})'\frac{\partial \tilde{\mathcal{M}}(f_{\omega_T}, D_{dev})}{\partial \omega_{t+1}}\\
=&(I - \eta_t \frac{\partial^2L_{\mu_\theta(s_t)}(f_{\omega_t},D_{train}^t)}{\partial \omega_t^2})'d\omega_{t+1}\\
=&d\omega_{t+1}-\eta_t \frac{\partial^2L_{\mu_\theta(s_t)}(f_{\omega_t},D_{train}^t)}{\partial \omega_t^2} d\omega_{t+1}.
\end{aligned}
\end{equation}

The last equation in Eqn. (\ref{eqn:dw_update}) leverages the symmetry of Hessian matrix: for a function $g(x,y)$, $\frac{\partial^2 g}{\partial x \partial y}=\frac{\partial^2g}{\partial y \partial x}$.

We further have the gradient of $\theta$ only incurred at timestep $t$ (i.e., via Eqn.(\ref{eqn:sgd_seq})), denoted as $d\theta|_t$, is:

\begin{equation}
\label{eqn:dtheta}
\begin{aligned}
d\theta|_t= \frac{\partial \tilde{\mathcal{M}}(f_{\omega_T}, D_{dev})}{\partial \theta}|_t&=(\frac{\partial \omega_{t+1}}{\partial \theta}|_t)'\frac{\partial \tilde{\mathcal{M}}(f_{\omega_T}, D_{dev})}{\partial \omega_{t+1}}\\
&= -\eta_t (\frac{\partial^2L_{\mu_\theta(s_t)}(f_{\omega_t}, D_{train}^t)}{\partial  \omega_t\partial\theta})'d\omega_{t+1}\\
&= -\eta_t \frac{\partial^2L_{\mu_\theta(s_t)}(f_{\omega_t}, D_{train}^t)}{\partial\theta\partial  \omega_t}d\omega_{t+1},
\end{aligned}
\end{equation}

where $\frac{\partial \omega_{t+1}}{\partial \theta}|_t$ represents the effect of $\theta$ to the value of $\omega_{t+1}$ happened only at timestep $t$, but not related with the effect to the value of $\omega_t$. Therefore we equivalently have $\frac{\partial \omega_{t}}{\partial \theta}=0$ in calculating $\frac{\partial \omega_{t+1}}{\partial \theta}|_t$. The last equation in Eqn. (\ref{eqn:dtheta}) again leverages the symmetry of Hessian matrix.  

By observing $d\theta=\sum_{t=0}^{T-1}d\theta|_t$, we obtain the recursive way to update $d\theta$ at timestep $t$ as in Eqn.(5) of the main paper:

\begin{equation}
d\theta = d\theta +d\theta|_t = d\theta-\eta_t \frac{\partial^2L_{\mu_\theta(s_t)}(f_{\omega_t}, D_{train}^t)}{\partial\theta\partial  \omega_t}d\omega_{t+1}.     
\end{equation}

\section{Experiment Details}
The details of network structures for the student models, the dataset used for neural machine translation, the training procedure for student and teacher models are provided here. 

\subsection{MNIST}
For MNIST dataset, we choose the simple multi-layer perceptron (MLP) and vanilla convolutional neural network (CNN) based LeNet architecture as our student models. 

The MLP contains only one single hidden layer with hidden size $500$, and the logistic regression output layer with size $10$. The input MNIST training sample is a flattened vector with size $28\times28$. The model is trained with mini-batch size $20$, momentum SGD \cite{momentum} is adopted with learning rate $0.01$ and momentum $0.9$ in straining the student model. 

The LeNet \cite{LeNet} model contains two (convolution + max-pooling) layers with kernel size $5\times5$ and filter number $20$, $50$ respectively, followed by one MLP with hiden size $500$. The model is trained with mini-batch size $500$ and the learning rate for momentum SGD update is $0.01$, the momentum is $0.9$.

\subsection{CIFAR-10/CIFAR-100}
For CIFAR-10 and CIFAR-100, we use the advanced CNN architecture ResNet \cite{ResNet} with different number of layers, and also the Wide-ResNet \cite{WRN}, DenseNet~\cite{DenseNet} which has superior performance.

We use the original and typical setting for the ResNet architecture. The inputs for the network are $32\times32$ images, with the per-pixel mean subtracted. The first layer is $3\times3$ convolutions, and then stack of $6n$ layers with $3\times3$ convolutions on the feature maps of sizes $\{32, 16, 8\}$. The numbers of filters are $\{16, 32, 64\}$ respectively. The subsampling is performed after convolutions with a stride of $2$. The network ends with a global average pooling layer, a 10-way (for CIFAR-10) or 100-way (for CIFAR-100) fully-connected layer and softmax layer. There are totally $6n+2$ stacked weighted layers. Identity shortcuts are connected to the pairs of $3\times3$ layers. We vary the $n=\{1,3,5\}$, leading to $\{8, 20, 32\}$-layer networks to evaluate our algorithm. The momentum optimizer with learning rate $0.1$ and momentum $0.9$ is conducted to update the student model, the learning rate is divided by $10$ after $40$ and $60$ epochs. The mini-batch size is $128$ in training. For data augmentation we do horizontal flips and take random crops from image padded by $4$ pixels on each side, filling missing pixels with reflections of original image. 

For CIFAR-10 dataset, we further adopt Wide-ResNet (WRN) and DenseNet as our student model. The WRN decreases the depth and increases witdth of ResNet. The specific configuration is WRN-$40$-$10$ setting, a ResNet with $40$ convolutional layers and a widening factor $10$ (the number of filters are $10$ times wider than the original ResNet, which is $\{160, 320, 640\}$). Other details are same as ResNet setting. For the DenseNet, the configuration is same as in \cite{DenseNet}, with bottleneck layers and compression module, named as DenseNet-BC. Specifically, the layer number $L$ is $190$ and the growth rate $k$ is $40$.

\subsection{IWSLT-14 German-English NMT}
For neural machine translation (NMT) experiment, the IWSLT-14 German-English \cite{IWSLT14} dataset we choose is a well-acknowledged benchmark in NMT literature. The training/dev/test dataset respectively contains roughly $153k/7k/7k$ sentence pairs. We process the German and English sentences to be $25k$ sub-word units by byte-pair-encoding (BPE) \cite{BPE} approach. The student model we used is based on LSTM \cite{lstm} with attention mechanism \cite{s2s_attn} and the Transformer \cite{transformer} network based on self-attention. The embedding size and hidden state size are both set as $256$. \emph{LSTM-1} contains only one LSTM layer while \emph{LSTM-2} has two LSTM hidden layers. Both student models are trained with simple SGD with learning rate $0.1$, the mini-batch size is $32$. The configuration for \emph{Transformer} is the $transformer\_small$ setting with $6$ layers of the encoder and decoder. To speed up the training process, as commonly done in previous works \cite{ac_s2s,rl4nmt}, we pretrain our student models for several epochs as warm-start models, and the training/dev set BLEU scores are computed based on the greedy searched translation results.

\subsection{Teacher Optimization}
For all experiments, the teacher models are optimized by Adam \cite{adam} with $\alpha=0.0001, \beta_1=0.9, \beta_2=0.999$ and $\epsilon=10^{-8}$. The teacher models are optimized with $60$, $100$ and $50$ steps (i.e., the number of teacher optimization steps in Algorithm 1 of the paper) for MNIST, CIFAR-10/CIFAR-100 and German-English translation tasks respectively. 

\bibliographystyle{plain}

\end{document}